\documentclass[10pt,twocolumn,letterpaper]{article}
\pdfoutput=1
\usepackage{iccv}
\usepackage{times}
\usepackage{epsfig}
\usepackage{graphicx}
\usepackage{amsmath}
\usepackage{amssymb}
\usepackage{todonotes}
\makeatletter
\let\@fnsymbol\@arabic
\makeatother

\usepackage[pagebackref=true,breaklinks=true,letterpaper=true,colorlinks,bookmarks=false]{hyperref}

\iccvfinalcopy 

\def\iccvPaperID{****} 

\ificcvfinal\fi

\begin{document}

\title{Scene Graph based Image Retrieval - A case study on the CLEVR Dataset}

\author{Sahana Ramnath\thanks{Indian Institute of Technology, Madras}\\
\and
Amrita Saha\thanks{IBM Research Labs, Bangalore}\\
\and 
Soumen Chakrabarti\thanks{Indian Institute of Technology, Bombay}\\
\and
Mitesh M. Khapra\footnotemark[1]\\
}

\maketitle
\ificcvfinal\thispagestyle{empty}\fi

\begin{abstract}
With the prolification of multimodal interaction in various domains, recently there has been much interest in text based image retrieval in the computer vision community. However most of the state of the art techniques model this problem in a purely neural way, which makes it difficult to incorporate pragmatic strategies in searching a large scale catalog especially when the search requirements are insufficient and the model needs to resort to an interactive retrieval process through multiple iterations of question-answering. Motivated by this, we propose a neural-symbolic approach for a one-shot retrieval of images from a large scale catalog, given the caption description. To facilitate this, we represent the catalog and caption as scene-graphs and model the retrieval task as a learnable graph matching problem, trained end-to-end with a REINFORCE algorithm. Further, we briefly describe an extension of this pipeline to an iterative retrieval framework, based on interactive questioning and answering.
\end{abstract}

\vspace{-3mm}
\section{Introduction}

With multimodal interaction becoming more common in domains like retail or travel, problems like text based image retrieval have picked up significant interest in the research community.  The challenges of such a problem involve understanding the nuances of interaction between the multiple modalities, and handling a real-time retrieval from large-scale catalog. While neural models with their rich expressibility to encode such complex modalities, have revolutionalized research on complex multimodal tasks, the standard practices of end-to-end pure neural style training fails to explicitly model the latent structures present in the different modalities or the different strategies required for the complex task. Without so, the neural model can make blatant mistakes, which its earlier symbolic counterparts would not have made.
In this work, we propose a neural symbolic approach for modeling a caption based image retrieval task. The backbone of such a modeling requires a scene-graph representation, \cite{johnson2015image} of the image catalog and the ongoing dialog context, and the retrieval task is modeled as a graph subsumption problem. This being one of the initial attempts at solving this problem in a neural-symbolic style, in this work, we showcase our proposed model on the popularly used diagnostic dataset of CLEVR \cite{johnson2017clevr} and CLEVR Dialog \cite{kottur2019clevr}. There have been some attempts at image retrieval on simpler versions of CLEVR-like environment, carefully created, where captions comprise of single tuples and there is no ambiguity in the image retrieval or where the dataset (for e.g. CSS in \cite{vo2019composing}) only considers three attributes and the task involves retrieving a target image given a reference image and some modification instructions in text \cite{vo2019composing}.

In general the task can be more complex, where the captions involve multiple interactions between many objects, and the catalog itself can be large scale and lastly, the caption can be too underspecified for retrieving a specific image. While, for the scope of this work, and due to space limitations, we focus on the one-shot retrieval, our broader motivation is towards an iterative approach where the retriever model can interactively ask further probing questions, and incorporate the feedback in its catalog search in order to quickly reach the target image. Our framework of scene-graph based image-retrieval allows for a neural symbolic approach to achieve this, by posing retrieval as a strategic search, and incorporating the strategies and structural constraints of the task in the neural reasoning  through the multimodal graphical representation. 

\vspace{-1mm}
\section{Proposed Model}
We now discuss a proposed model that can incorporate these aspects pragmatically for retrieving images from a catalog given a caption description.


\textbf{Scene graph based parsing of the Catalog images:} We first pretrain a scene-graph generation model in the CLEVR setting, which involves
obtaining a symbolic scene graph corresponding to each catalog image before training the retrieval model. In such a symbolic scene graph, nodes correspond to bounding boxes in the image, and are labeled with a set of attributes and edges define spatial relations between objects. The nodes can be represented by an $M \times N$ matrix, where $M$ is the number of attributes (like color, shape) and $N$ is the dimension of the Glove based \cite{pennington2014glove} semantic features or the visual features of the attribute label, and edge is represented by the GloVe embedding of the label.
\\
\textbf{Catalog Scene Graph:} We next represent the entire catalog of images as a single consolidated scene-graph, which we call the `\emph{catalog graph'}. The set of vertices for this graph is the union of the sets of vertices over all the images in the catalog. Note that since a node is identified by its attributes, identical nodes from different images will collapse into a single node in the consolidated scene graph representation. Similarly, the incoming or outgoing edges for each node in the catalog scene graph will be the union of the incoming or outgoing edges for that node, over all images the node appears in. 
Further every node and edge of this catalog graph will be associated with an inverted index of images, which contain that node or edge. 
\\
\textbf{Caption Scene Graph:} We represent the input caption as a scene graph, using the SPICE framework\cite{spice2016}, which uses the dependency tree structure of the sentence, followed by multiple tree transformations to get the final precise scene graph representation of the text. This graph, which we define as \emph{`query graph'} uses the GloVe based semantic embeddings for the attribute-values describing an object as its node embedding and that of the relation label as the edge embedding. Note that some of the attributes of the objects in the query graph may be unknown, for which we use zero embedding, to get a $M \times N$ matrix representation. 
\\
\textbf{One-shot Image-Retrieval:} We now discuss the pipeline for a one-shot retrieval, given a query graph and the catalog scene-graph. For that, we need to map the query graph to a minimal subgraph of the catalog-graph which subsumes the query graph. Graph G1 is subsumed by another graph G2, if every node and head,edge,tail of G1 is subsumed by some node or triple in G2. We define node subsumption score between a query node and a catalog node as an inverse of the weighted version of the Frobenius distance between the node representations. The weight vector is a $M$-dimensional boolean vector representing whether the corresponding attribute-value is mentioned in the query node or not, making the distance metric assymetrical, and scoring all catalog nodes that match the query in terms of \emph{only} the \emph{known} attributes, higher. 
Similarly a subsumption scoring function across head,edge,tail triples can be defined as the summation of the head and tail node subsumptions along with that of the edge type. Using this, for every node and triple in the query graph, we construct an attention map over the nodes and triples in the catalog graph. 
\\
\textbf{Action Space and Sampling:} We define our action space for the retrieval as set of catalog images; the action sampling probability is the probability of sampling the exact subgraph of the catalog graph that corresponds to the image's scene graph, as the graph subsuming the query graph. For that we first compute, for each node or each head-edge-tail triple in the query graph, the \emph{best-match} score (by taking a \emph{max} over the attention map scores) over the nodes or triples in the image's scene graph. The final probability assigned to the image is the normalized product of these scores received by all the nodes and triplets in the query graph.
\\
\textbf{Training}: Using this action sampling mechanism, the pipeline is trained using the REINFORCE \cite{sutton2000policy} algorithm
\vspace{-2mm}

\begin{align*}
\boxed{
 J(\theta)  = \sum_{images} P_{\theta}(\text{image})[R_{\theta}(\text{image})-B] \log(P_{\theta}(\text{image}))
 }
\end{align*}

\vspace{-1.5mm}

Here $\theta$ represents the network's parameters, $P_{\theta}(\text{image})$ is the action probability calculated above, $R_{\theta}(\text{image})=1-normalized(L2(I-I'))$ where $I$ is the representation of the gold image and $I'$ is that of the retrieved one. The framework allows for other reward functions based on ranking loss or rank of the gold image in the retrieved list. As a sanity check we tested our one-shot image retrieval model, with the query graph being randomly sampled partial-scene graphs of images from the CLEVR dataset. Starting with a retrieval accuracy of $99.9\%$ when the full scene-graph is used, it drops to $89\%$ and subsequently $75\%$ when $20\%$ and $30\%$ of the nodes are dropped from the query.
\\
\textbf{Extension to Iterative Retrieval:} We now give a brief motivation on how to build up on an iterative retrieval process, by modeling a questioner and answerer with this one-shot retrieval pipeline as backbone. At each iteration of the dialog starting from the caption, the Questioner model uses its partial filled query graph to do a preliminary image retrieval using the one-shot pipeline. Based on the subgraph of the catalog graph, corresponding to the $K$ most similar images, the questioner defines an action space, where each action is a potential question that can be asked on a node or triple, whose corresponding query-side counterpart has missing information. These actions can be sampled on heuristically guided learnable policies, based on how much information an action adds to the current query, or how well the action shatters the catalog search space, so that the target image can be reached in fewest iterations. Simultaneously, the Answerer is trained as a Scene-graph based VQA model. The answer is then incorporated by the Questioner, which models it as a problem of learning to update the node and edge embeddings of the query graph, based on the feedback. This iteration continues till the model decides to terminate and retrieve the closest image.
\vspace{-1mm}
\section{Conclusion}  
We introduced a scene graph based image retrieval framework which models the task as a learnable graph matching problem over the query and catalog graph. We further also motivated how this neural-symbolic technique extends to an iterative retrieval framework, by stategizing multiple rounds of questioning and answering between two jointly trained models, to reach the target.

{\small
\bibliographystyle{ieee_fullname}
\bibliography{egbib}
}

\end{document}